\documentclass{article}

\usepackage[final]{iai_neurips_2024}

\usepackage[T1]{fontenc}
\usepackage{graphicx}
\usepackage{booktabs}
\usepackage[misc]{ifsym}

\usepackage{amsmath,amssymb}

\usepackage[utf8]{inputenc} 

\usepackage{hyperref}       
\usepackage{url}            
\usepackage{adjustbox}
\usepackage{amsfonts}       
\usepackage{nicefrac}       
\usepackage{microtype}      
\usepackage{xcolor}         

\usepackage{float}

\usepackage{enumitem}

\title{The effect of whitening\\ on explanation performance}

\author{%
  \textbf{Benedict Clark}$^{1}$ \quad
  Stoyan Karastoyanov$^{2}$ \quad
   Rick Wilming$^{1,2}$ \quad
   \textbf{Stefan Haufe}$^{1,2,3}$ \quad\\
   $^1$Physikalisch-Technische Bundesanstalt, Berlin, Germany \\
   $^2$Technische Universität Berlin, Germany \\
   $^3$Charité – Universitätsmedizin, Berlin, Germany
 }

\begin{document}

\maketitle

\begin{abstract}
Explainable Artificial Intelligence (XAI) aims to provide transparent insights into machine learning models, yet the reliability of many feature attribution methods remains a critical challenge.
Prior research \citep{haufe_interpretation_2014,wilming_scrutinizing_2022,wilming_theoretical_2023} has demonstrated that these methods often erroneously assign significant importance to non-informative variables, such as suppressor variables, leading to fundamental misinterpretations.
Since statistical suppression is induced by feature dependencies, this study investigates whether data whitening, a common preprocessing technique for decorrelation, can mitigate such errors.
Using the established XAI-TRIS benchmark \citep{clark_xai_2024}, which offers synthetic ground-truth data and quantitative measures of explanation correctness, we empirically evaluate 16 popular feature attribution methods applied in combination with 5 distinct whitening transforms.
Additionally, we analyze a minimal linear two-dimensional classification problem \citep{wilming_theoretical_2023} to theoretically assess whether whitening can remove the impact of suppressor features from Bayes-optimal models.
Our results indicate that, while specific whitening techniques can improve explanation performance, the degree of improvement varies substantially across XAI methods and model architectures.
These findings highlight the complex relationship between data non-linearities, preprocessing quality, and attribution fidelity, underscoring the vital role of pre-processing techniques in enhancing model interpretability.
\end{abstract}

\section{Introduction}
In recent years, there has been a growing focus on empirically validating the performance of so-called explainable artificial intelligence (XAI) methods by examining the accuracy of their explanations  \citep[such as, ][]{tjoa_quantifying_2020, li_experimental_2021, zhou_feature_2022, arras_clevr-xai_2022, gevaert_evaluating_2022, agarwal_openxai_2022, oliveira_benchmarking_2024, wilming_gecobench_2024}. 
While some such studies use ground-truth explanations, they often face limitations in their objective assessment of explanation correctness, the variety of XAI methods analyzed, and the complexity of the explanation tasks. 
Many existing ground-truth problems are designed in a way that avoids realistic correlations between class-related and class-unrelated features (such as image foreground versus background). 
In real-world scenarios, however, such dependencies can introduce suppressor variables, noisy features that are not directly associated with the prediction target but can be utilized by the model \citep[for example, for denoising, e.g.,][]{haufe_interpretation_2014}. 
For instance, in image data, background elements representing lighting conditions could act as suppressor variables. 
A model may leverage this information to adjust for lighting variations, thereby enhancing object detection. 
More comprehensive discussions on suppressor variables are available in \citet{conger_revised_1974, friedman_graphical_2005, haufe_interpretation_2014, wilming_theoretical_2023}.

A common XAI paradigm is to assign an `importance' score to each feature of a given input. It has been shown, though, empirically and theoretically, that various popular feature attribution methods tend to systematically assign importance to suppressor variables even in linear settings \citep{wilming_scrutinizing_2022,wilming_theoretical_2023}. 
Extending this result, the work of \citet{clark_xai_2024} introduces the XAI-TRIS datasets, composed of four binary image classification problems, one linear and three non-linear. 
In each dataset, different types and combinations of tetrominoes \citep{golomb_polyominoes_1996}, geometric shapes consisting of four blocks, need to be distinguished from one another. 
These tetromino images are overlaid on different types of noisy backgrounds: white noise (WHITE) and correlated (CORR) background; the latter induces a suppression effect through Gaussian smoothing.

The tetrominoes then represent discriminative features serving as ground truth explanations.
\citet{clark_xai_2024} show that contemporary XAI methods fail to highlight tetrominoes consistently and, in some cases, are outperformed by model-ignorant edge detectors. 

It is assumed that the suppression effect degrades explanation performance, and one potential approach to reduce this impact is to use data whitening techniques. 
Whitenings are multivariate linear transformations that transform the original features into a new space in which all features are uncorrelated and have unit variance, thus reducing feature redundancy. Notably, some whitening transformation maintain a 1:1 correspondence between original and transformed features, making it possible to visualize importance attributions in input space and assessing their efficacy as explanations.


In this paper, we take the XAI-TRIS datasets and use the associated experimental pipeline proposed by \citet{clark_xai_2024} to assess whether the use of whitening techniques can improve the performance of XAI methods with respect to correctness of the explanations produced.
The data scenario where the WHITE background type gets utilized serves as a baseline due to having no correlations between features of the background, hence we are not applying whitening methods.
Then we test if applying whitening methods to the CORR background type can reduce the impact of suppressor variables. Here, we hypothesize explanations to be more aligned with discriminative features, and hence to see improved explanation performance.
Finally, also take the existing theoretical framework proposed by  \citet{wilming_theoretical_2023}, in order to analyze whether whitening methods can analytically remove the impact of a suppressor feature in a two-dimensional classification problem.
%
%
\section{Methods}
Our general workflow of applying and benchmarking post-hoc XAI methods follows previous work \citep{wilming_scrutinizing_2022, clark_xai_2024, clark_exact_2024}.
We take a dataset generated with explicitly known class-related features defining the classification task and the ground truth for explanations, and train a machine learning model.
The trained model is then applied to test inputs, for which output explanations are computed by XAI methods. We exclusively consider feature attribution methods, which assign an `importance' score to each feature of the input. 
We then apply two performance metrics to compare produced explanations and the ground truth explanation for the given sample, giving us measures of the explanation performance of each method.
Below, we highlight each of these steps, with more depth and the exact parameterizations given in the appendices.

\subsection{Data Generation}\label{subsec:datagen}
We utilize the datasets supplied by the XAI-TRIS suite \citep{clark_xai_2024}, providing four binary image classification problems.
We make use of the $8 \times 8$-px variant with two background types -- the uncorrelated (WHITE) and correlated (CORR) backgrounds.
The CORR background type takes the WHITE background and smoothes it with a Gaussian filter, inducing correlations between features. 
This also induces a suppression effect where background pixels overlapping with the placed tetromino are correlated to nearby background pixels.
The data generation process is described in full detail in Supplementary materials Section \ref{app:data-gen}, however briefly, the four classification scenarios are defined as:
\begin{enumerate}
    \item \textbf{Linear (LIN)} In the linear case, the classification problem is between a T-shaped tetromino versus an L-shaped tetromino pattern placed at the same fixed positions of the image throughout the entire dataset.

    \item \textbf{Multiplicative (MULT)} The multiplicative scenario is similar to the LIN scenario with classifying T- versus L-shaped tetrominoes, however here each tetromino is multiplied with the background to induce non-linearity.
    
    \item \textbf{Translations and rotations (RIGID)} Here, the T- and L-shaped tetromino patterns are not in a fixed position and are randomly translated and rotated anywhere in the sample, and added together with the underlying background.
    
    \item \textbf{XOR} Both of the T- and L- shaped tetrominoes are present in each sample, but the classification problem is defined as the XOR configurations of adding or subtracting both tetrominoes from the background versus adding/subtracting one of each tetromino in the other class.
\end{enumerate}
%


\subsubsection{Data whitening methods}\label{subsec:whitening}
Data whitening is achieved by applying a linear transformation that adjusts the direction and scale of the data. 
The process typically involves eigenvalue decomposition of the covariance matrix and normalizing the eigenvalues \citep{kessy_optimal_2018}. 
We study the following five techniques for which the supplementary materials Section \ref{app:whitening} contains more details.

\paragraph{Eigendecomposition-based Whitening (Sphering, Symmetric Orthogonalization, OSP Whitening)} 
These methods whiten data via an eigendecomposition of a matrix $\mathbf{M}$, followed by rescaling to produce uncorrelated features. The difference lies in the choice of matrix $\mathbf{M}$:
\begin{itemize}
    \item \textit{Sphering}: $\mathbf{M}$ is the sample covariance matrix. This method transforms the data by projecting it onto its principal axes and rescaling to unit variance along each axis, followed by a rotation back into input space. The resulting features are uncorrelated and standardized, with a one-to-one correspondence to the original features \citep{kessy_optimal_2018}.
    
    \item \textit{Symmetric Orthogonalization}: $\mathbf{M}$ is the overlap matrix, i.e., the (un-centered) second-moment matrix of the data \citep{annavarapu_singular_2013}. The goal is to produce mutually uncorrelated features while minimizing the difference between original and transformed features in a least-squares sense. When the data is centered, the overlap matrix is equivalent to the covariance matrix.
    
    \item \textit{Optimal Signal Preservation (OSP) Whitening}: $\mathbf{M}$ is the correlation matrix. This method emphasizes preserving the signal structure of individual features while removing redundancy. It follows the same whitening procedure but uses the correlation matrix instead of the covariance or overlap matrix \citep{kessy_optimal_2018}.
\end{itemize}




\paragraph{Cholesky Whitening} Cholesky whitening \citep{kessy_optimal_2018} applies the Cholesky decomposition to the covariance matrix. 
This decomposition leads to a lower triangular transformation matrix that leads to uncorrelated uniform-variance transformed features. 
Notably, the triangular structure induces an ordering, whereby the first feature remains unchanged, the second feature gets orthogonalized w.r.t. the first, the third feature gets orthogonalized w.r.t to the first two, and so on. 
Hence, this whitening depends on the order of pixels, which in our case is $(H, W)$ for H, the height of the image and W, the width.
The top-left pixel in the image remains unchanged, with the subsequent orthogonalization following horizontally across each row of pixels. 

\paragraph{Partial Regression} In contrast to the global approaches of the previous methods, partial regression \citep{velleman_regression_1981} focuses on removing the linear dependence of each feature on the others, one at a time. This approach involves regressing each feature against all others and replacing it with the residuals of this regression. While this method does not necessarily ensure uncorrelated features with unit variance, it aims to remove some of the shared information between them.

For each of these whitening and related techniques, the XAI-TRIS data is initially centered, and the resulting covariance matrix (when used) is regularized to ensure numerical stability.
This is done by adding a value slightly larger than the absolute minimum eigenvalue to the diagonal of the covariance matrix, if the smallest eigenvalue is negative or very close to zero.
Here, we compare the minimum eigenvalue to the threshold $1 \times 10^{-16}$.

\subsection{Classifiers}\label{subsec:classifiers}

Following the approach of \citet{clark_xai_2024}, three different architectures are employed: (1) a Linear Logistic Regression (LLR) model, a single-layer fully-connected neural network; (2) a Multi-Layer Perceptron (MLP) with four fully-connected layers, using ReLU activations; and (3) a Convolutional Neural Network (CNN) with four ReLU-activated convolutional layers followed by max-pooling.
All models lead to a two-neuron softmax-activated output layer.
We train a model for each CORR scenario and also for each whitening method applied to the respective CORR scenario, making use of the appropriate whitened data for each model.
We ensure that each trained model achieves at least $80\%$ test accuracy, so that the resulting trained models have comparative performance.
Specific details are described in supplementary materials Section \ref{app:classifiers}.

\subsection{XAI Methods}
We analyze sixteen widely recognized methods within the domain of XAI.
The core discussion is centered around the evaluation of four distinct XAI methods: Local Interpretable Model-Agnostic Explanations (LIME) \citep{ribeiro__2016}, Layer-wise Relevance Propagation (LRP) \citep{bach_pixel-wise_2015}, Gradient SHAP \citep{lundberg_unified_2017}, and Integrated Gradients \citep{sundararajan_axiomatic_2017}. 
The full list of methods studied and the associated results can be seen in supplementary materials Section \ref{app:xai-methods}.
Predominantly, default parameters are adhered to, with exceptions noted where a baseline $b=0$ is explicitly defined, reflecting a widely recognized convention in the field \citep{mamalakis_carefully_2022}. 

The input for an XAI method is a trained ML model, the given test sample or batch of multiple samples designated for explanation, and (where relevant) the baseline test reference $b = 0$. 
The full results presented in the appendices (Figures \ref{fig:methodsEMD} and \ref{fig:methodsPrecision}) make use of four model-ignorant techniques to establish baselines of explanation performance.
This enables the assessment of whether the often intricate XAI methods genuinely offer superior explanations compared to approaches devoid of model-specific insights. 
The first method considered is the Sobel filter, employing both horizontal and vertical filter kernels to estimate the first-order derivatives of data. 
The second method utilized is the Laplace filter, which approximates the second-order derivatives of data using a single symmetrical kernel. 
Both methodologies serve as edge detection operators and are applied to each test sample. Additionally, random samples from a uniform distribution and the rectified test data sample itself are employed as `explanations' for comparison purposes.

\subsection{Explanation Performance Metrics}\label{subsec:metrics}
We define a `correct' explanation as one which highlights truly important features for the classification task (i.e. any subset of features forming the tetrominoes) and does not place false-positive importance on features outside of said ground truth.
We adopt the quantitative metrics used by \citet{clark_xai_2024}, namely precision and earth mover's distance (EMD).
These metrics serve as an objective and empirical foundation for analyzing how well a model's explanations align with a set of class-dependent features identified as ground truth. 

The precision metric is calculated as the ratio of the correctly identified features within the top $k$ features ranked by their absolute importance scores to the total number of truly important features identified in the sample.
The focus on the highest-ranking features reflects the real-world scenario where only the most influential factors are typically considered in decision-making processes (e.g., a doctor using a subset of symptoms to form a diagnosis).

The EMD quantifies the minimal expenditure required to transform one distribution into another. It is also known as the optimal transport distance.
Applied in our context, this involves the cost needed to transform a continuous-valued explanation into the ground truth, with both distributions normalized to have equal `mass'.
The calculation of EMD utilizes the Euclidean distance between pixels as the ground metric.
To calculate the EMD, we use the algorithm introduced by \citet{bonneel_displacement_2011} as implemented in the Python Optimal Transport library by \citet{flamary_pot_2021}. 
A normalized EMD performance score is defined by taking the optimal transport from an explanation to ground truth and dividing by the maximum euclidean distance possible.
In practice we take one minus this score such that a score of 1.0 is the `perfect' explanation.

As discussed by \citet{clark_xai_2024}, both metrics assess the model's ability to highlight features that are truly relevant, as per the ground truth, while minimizing the inclusion of less significant (false-positive) features. 
\section{Results}
\subsection{Qualitative Analysis}
%
Supplementary Figure \ref{fig:Gheatmaps} depicts the absolute-valued global importance heatmaps, the mean of all explanations for every correctly-predicted sample, for the LIN, MULT, and XOR scenarios. 
As the RIGID scenario has no static ground truth pattern, calculating a global importance map is not possible. 
Shown are results for four XAI methods (Gradient SHAP, LIME, LRP, and Integrated Gradients respectively) for each of the three models (LLR, MLP, CNN respectively) followed by the model-ignorant Laplace filter. 
This is shown for a random correctly-predicted sample, including the RIGID scenario,  in supplementary Figure \ref{fig:randomSample}.

Interestingly, not all whitening techniques impact XAI interpretations uniformly. 
The Cholesky whitening and partial regression techniques demonstrate slightly more focused attributions but still show notable activity in regions outside the foreground signal. 
This indicates that while some varied amount of suppression of background noise is achieved, overall the techniques seem more prone to the negative influence of suppressor variables on explanation performance. 
Contrasting that, the optimal signal preservation and sphering methods, designed to preserve more of the data structure, only subtly modify explanations and can be observed to yield importance maps that more closely aligned with the true signal, indicating a stronger reduction of potential suppressor variable influence. 
Symmetric orthogonalization presents the most concentrated patterns of importance, closely mirroring the ground truth and demonstrating the highest resilience to the potentially misleading effects of suppressor variables among the examined whitening techniques. 
Supplementary Figure~\ref{fig:randomSample} presents the importance maps obtained for a correctly-predicted data sample, for data with no whitening applied and data for which the symmetric orthogonalization whitening method was applied.
Within the variety of XAI methods, gradient-based methods like Gradient SHAP and Integrated Gradients illustrate an observable evolution from more dispersed attribution patterns in the non-whitened case, to more concentrated patterns as the data undergoes the various types of whitening. 

\subsection{Quantitative Analysis}
%
\begin{figure}[!ht]
    \includegraphics[width=\textwidth]{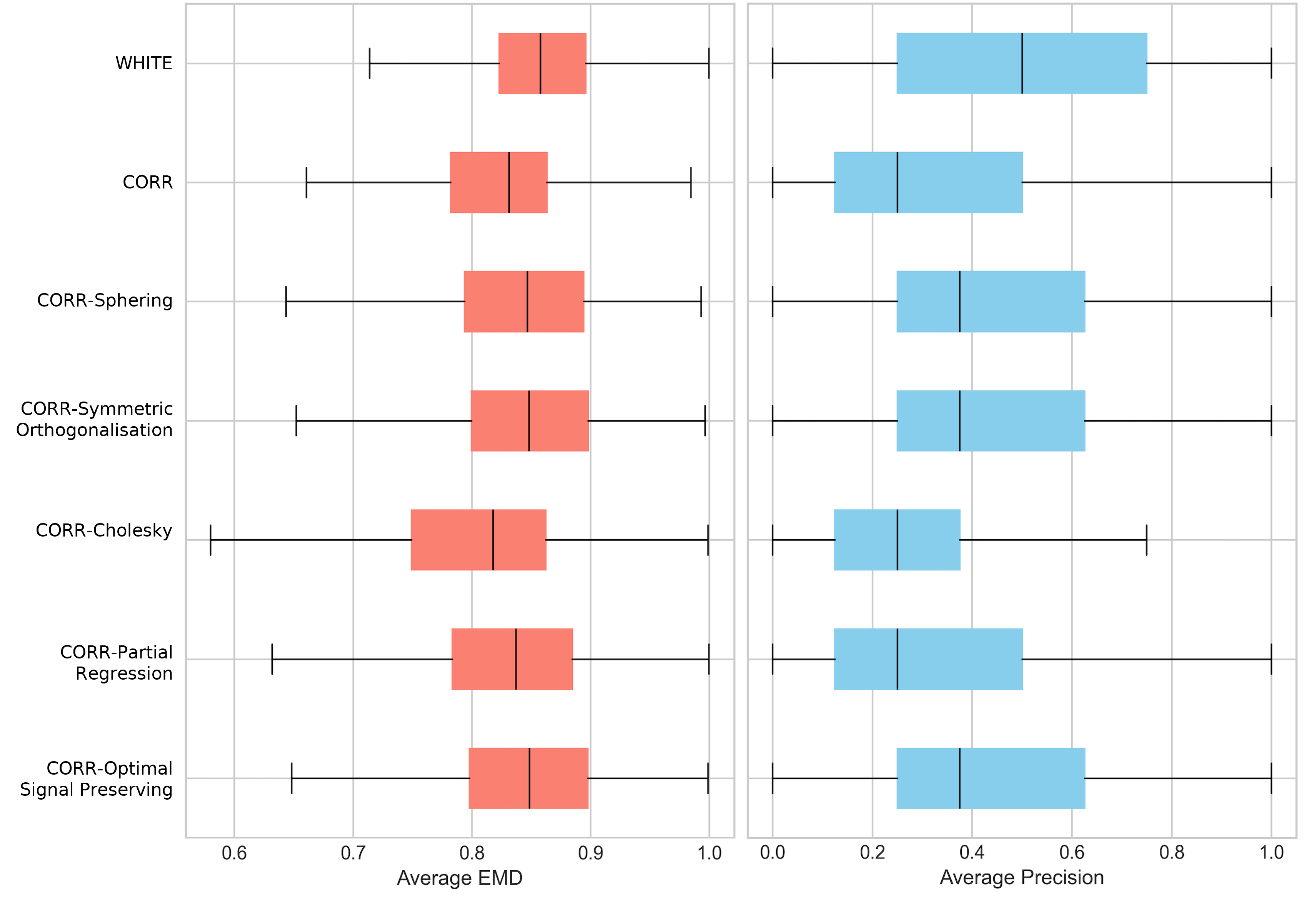}
    \caption{Average earth mover's distance (left) and precision (right) across all samples, XAI methods, and XAI-TRIS scenarios. This is split by background type, where the WHITE background serves as a `baseline' with the CORR background type serving as the base for whitening. Each subsequent row therefore shows the application of different whitening methods to the underlying CORR background scenarios. Both metrics follow a similar trend of which whitening methods improve the correctness, where the sphering, symmetric orthogonalization, and optimal signal preserving methods perform the best -- nearly reaching the performance levels of the WHITE results.}\vspace{-1.5ex}
    \label{fig:quantitative-results}
\end{figure}
%
%
From Figure \ref{fig:quantitative-results}, it can be observed that the precision of XAI methods tends to decline when operating on the correlated background (compared to the uncorrelated case), confirming the notion that correlated noise negatively impacts the ability of XAI methods to accurately identify features of importance, due to the induction of suppressor variables. 
This decline is rectified to varying extents by the application of whitening techniques, which aim to remove the correlation between features in the dataset, thereby mitigating these potential suppressor variables that could lead to false attributions of importance. 
In this regard, Sphering, Optimal Signal Preserving, and Symmetric Orthogonalization stand out as the most effective techniques in restoring precision, indicating their strength in clarifying the data's structure and enhancing the ability of XAI methods to discern true signals from noise. 
As with the precision metric, sphering, symmetric orthogonalization and the optimal signal preserving transformation demonstrate the highest EMD values.
Formulated on similar principles with the main difference being the choice of matrix to be diagonalized in the eigenvalue decomposition, symmetric orthogonalization (diagonalizing the overlap matrix) and optimal signal preservation (diagonalizing the correlation matrix) have near identical results. 
Only the lower quartile EMD performance for symmetric orthogonalization looks to be slightly higher than for optimal signal preservation.
For both metrics, it can be observed that partial regression and Cholesky whitening, while sometimes improving upon the correlated scenario, fall short compared to the other techniques. 
This suggests that while they do have a positive impact, they might not be as capable of dealing with complex correlations or might introduce artifacts that prevent the XAI methods from reaching the accuracy levels of the other techniques. 

Supplementary Figures \ref{fig:EMDFour} and \ref{fig:precisionFour} expand on the results of Figure \ref{fig:quantitative-results} by splitting up results for each problem scenario and background type, and by the four main XAI methods studied. Supplementary Figures \ref{fig:methodsEMD} and \ref{fig:methodsPrecision} go even further by illustrating the EMD and precision results for all sixteen XAI methods studied and four baselines for the non-whitened case, compared to the top performing whitening technique as identified by the qualitative analysis -- symmetric orthogonalization. 
While we can see improvement in explanation performance in many cases where whitening is used, the results are not consistent across all XAI methods.
%
\subsection{Theoretical analysis}
To relate these results to theory, we adopt the two-dimensional linear generative model introduced by \citet{wilming_theoretical_2023}, where $\mathbf{x} = \mathbf{a} z + \eta$ and $y = z$, with $Z \sim Rademacher(1/2)$ the random variable of the realization $z$, signal pattern $\mathbf{a}=(1, 0)^{\top}$ and $H \sim N(\mathbf{0}, \Sigma_\eta)$ the random variable of the realization $\eta$. 
The noise covariance and subsequent data covariance matrices are parameterized as follows:
\begin{equation}
    \label{eq:noise-covariance}
    \begin{split}
        \Sigma_\eta = \begin{bmatrix}
                 s_1^2 & c s_1 s_2 \\ c s_1 s_2 & s_2^2 \;,
        \end{bmatrix} \;,
        \quad \Sigma_\mathbf{x} = \begin{bmatrix}
                 s_1^2 + 1& c s_1 s_2 \\ c s_1 s_2 & s_2^2
        \end{bmatrix} \;,
    \end{split}
\end{equation}
where $s_1$ and $s_2$ are non-negative standard deviations of $x_1$ and $x_2$ respectively and $c \in [-1, 1]$ is the correlation coefficient.

This defines a two-dimensional feature space where $x_1$ is the truly important variable, statistically related to target $y$, and $x_2$ is an explicit suppressor variable. 
\citet{wilming_theoretical_2023} derive the Bayes-optimal linear classifier $f(x) = w^{\top}x + b$ with weights $w^{\top}=(w_1, w_2)^{\top} = \Sigma^{-1} (\mu_1 - \mu_2)$ for class means $\mu_1 = (1,0)^{\top}$ and $\mu_2 = (1,0)^{\top}$. 
The classification problem is set up such that no offset is required, so $b=0$.
The authors show that, when $|c|>0$, the Bayes-optimal model has to place non-zero weight on $w_2$, relating to suppressor $x_2$, despite $x_2$ having no class-related information.
Traditionally, people interpret linear models (and consider them intrinsically interpretable or 'white box' models) by the strength of the model weights, where a feature with significantly non-zero weight is deemed important \citep{haufe_interpretation_2014}. Therefore, in this case, the Bayes-optimal model leads to the suppressor being considered important despite containing no class-related information.
\citet{wilming_theoretical_2023} go on to show that this  impacts the explanations of popular feature attribution methods in the sense that the non-zero model weight on $x_2$ carries over to nonzero attribution to $x_2$. 
\begin{figure}[!ht]
    \includegraphics[width=\textwidth]{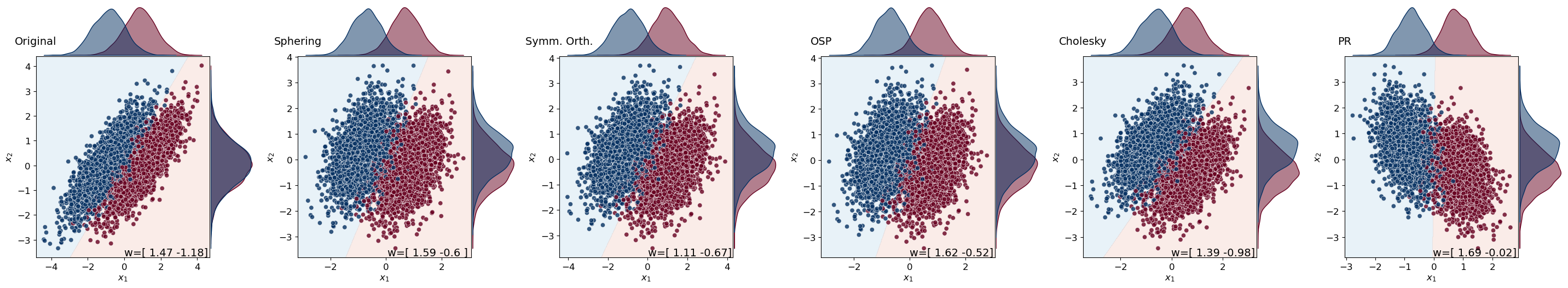}
    \caption{Data sampled from the linear generative process of the 2D-suppressor data \citep{wilming_theoretical_2023}, with correlation coefficient $c=0.8$ and variances $s_1^2=1.0$ and $s_2^2=1.0$. Boundaries of the Bayes-optimal decision are shown as well, where the marginal distribution of suppressor variable $x_2$ for the original data shows that it does not carry any class-related information. Only Partial Regression is able to remove the influence of the suppressor from the Bayes-optimal decision.}
    \label{fig:2d_whitening}
\end{figure}
Here, we apply each whitening method to the same setup to see if the whitening methods can analytically remove the suppression effect from this two-dimensional data, via the strength of the resulting weight vectors.
Figure \ref{fig:2d_whitening} shows the two-dimensional suppressor data followed by the application of each whitening method to these data. 
Visually, only Partial Regression is able to correct the Bayes-optimal decision rule in this setting. 
Mathematically, once we apply the whitening, we calculate the inverse of the whitened covariance matrix and then derive the weights via $w^{\top}= \Sigma^{-1} (\mathbf{W}\mu_1 - \mathbf{W}\mu_2)$ for whitening matrix $\mathbf{W}$.
The full methodology for calculating the weight vector of the Bayes-optimal model after each whitening transformation is detailed in the supplementary materials.

\vspace{-0.4ex}Starting with the eigendecomposition-based methods, the weights of the Bayes-optimal model after sphering are mathematically somewhat involved, though clearly $w_2$ being non-zero shows that the suppressor is not able to be removed by whitening. With $\alpha = \sqrt{4s_1^2 s_2^2 c^2 + (s_1^2 - s_2^2 + 1)^2}$, 
$\beta = s_1^2 + s_2^2 + 1$, and 
$\gamma = \left(s_1^2(c^2 - 1) - 1\right)\sqrt{8s_1^2s_2^2c^2 + 2(s_1^2 - s_2^2 + 1)^2}$,\vspace{-0.9ex}
\begin{align*}
w_1^{sphering} &= 
\frac{
  -\left( (s_1^2(2c^2 - 1) + s_2^2) \left( \sqrt{-\alpha + \beta} - \sqrt{\alpha + \beta} \right) \right)
}{\gamma} \nonumber \\
&\quad + \frac{
  \sqrt{-\alpha + \beta}
  - \sqrt{\alpha^2(-\alpha + \beta)}
  - \sqrt{\alpha + \beta}
  - \sqrt{\alpha^2(\alpha + \beta)},
}{\gamma} \\
w_2^{sphering} &= 
\frac{
  s_1 c \left(
    (s_1^2 + s_2^2) \left( \sqrt{-\alpha + \beta} - \sqrt{\alpha + \beta} \right)
    + \sqrt{-\alpha + \beta}
  \right)
}{s_2 \gamma} \nonumber \\
&\quad + \frac{
  s_1 c \left(
    + \sqrt{-\alpha^2(\alpha - \beta)}
    - \sqrt{\alpha + \beta}
    + \sqrt{\alpha^2(\alpha + \beta)}
  \right)
}{s_2 \gamma}.
\end{align*}
For symmetric orthogonalization, our implementation also scales by the inverse of the features' standard deviations. 
While this produces fundamentally the same transformation as doing a symmetric orthogonalization in the exact way as sphering (but with the overlap matrix), the calculations produce a different whitening matrix, and therefore different weights for the resulting Bayes-optimal model. 
For the XAI-TRIS data this has no impact, as the data is re-scaled to the range $[-1,1]$ after each whitening transformation.

Here, we show the result for symmetric orthogonalization as implemented experimentally, with the supplementary materials showing the alternative approach that appears mathematically closer in line to sphering and OSP. 
There, we also show that the two transformations are visually equivalent. 
Here, we also have the data sample size N in the resulting weight vector, as outlined in supplementary Section \ref{app:theory}.
We substitute $\alpha = \sqrt{4 a^2 (a^2 + 1) b^4 c^2 + \left((a^2 + 1)^2 - b^4\right)^2}$, 
$\beta = (N - 1)^2 a^4 + 2 (N - 1)^2 a^2 + (1 + b^4)(N - 1)^2$, and 
$\gamma = a^2(c^2 - 1) - 1$, and form the weight vector $w$ of the Bayes-optimal model as
\begin{align*}
w_1^{sym} &= -\frac{4 \sqrt{2} (1 + a^2) b^4 \cdot \gamma \cdot (N - 1)^5}{(\beta - \alpha)^{3/2} \cdot (\beta + \alpha)^{3/2} \cdot \sqrt{\alpha}} \cdot \Bigg[ \\
&\quad \left( \sqrt{\beta - \alpha} - \sqrt{\beta + \alpha} \right)
\left( a^4 (N - 1)^2 + 2 a^2 (N - 1)^2 - b^4 (N - 1)^2 \right) \\
&\quad + \alpha \left( \sqrt{\beta - \alpha} + \sqrt{\beta + \alpha} \right) - 2 N \left( \sqrt{\beta - \alpha} - \sqrt{\beta + \alpha} \right) \\
&\quad + N^2 \left( \sqrt{\beta - \alpha} - \sqrt{\beta + \alpha} \right) + \sqrt{\alpha} \left( \sqrt{\beta - \alpha} + \sqrt{\beta + \alpha} \right)
\Bigg] \\[2ex]
w_2^{sym} &= -\frac{8 \sqrt{2} a (1 + a^2)^2 b^5 c \cdot \gamma \cdot (N - 1)^7 \cdot \left( \sqrt{\beta - \alpha} - \sqrt{\beta + \alpha} \right)}{(\beta - \alpha)^{3/2} \cdot \left( \beta + \sqrt{\alpha} \right)^{3/2} \cdot \sqrt{\alpha}}.
\end{align*}
For the final of the eigendecomposition whitening methods, Optimal Signal Preserving (OSP), we receive a relatively simpler weight vector, but still with non-zero weight on suppressor feature $w_2$. With $\alpha = \sqrt{-\sqrt{a^2+1}ac+a^2+1}$ and $\beta = \sqrt{\sqrt{a^2+1}ac+a^2+1}$,
\begin{align*}
    w_1^{OSP} &= \frac{ac\left(\sqrt{1-\frac{ac}{\sqrt{a^2+1}}}-\sqrt{\frac{ac}{\sqrt{a^2+1}}+1}\right) + \alpha + \beta}{1-a^2\left(c^2-1\right)}, \\
    w_2^{OSP} &= \frac{ac\sqrt{1-\frac{ac}{\sqrt{a^2+1}}}+ac\sqrt{\frac{ac}{\sqrt{a^2+1}}+1}+\alpha-\beta}{a^2\left(c^2-1\right)-1}.
\end{align*}
For Cholesky whitening, the Bayes-optimal weights are
\begin{align*}
    w_1^{Chol} = \frac{2}{\sqrt{a^2+1}}, \quad
    w_2^{Chol} = -\frac{2ac}{\sqrt{\left(a^2+1\right)\left(1-a^2\left(c^2-1\right)\right)}}.
\end{align*}
%


Finally, we analytically back up the result of Figure \ref{fig:2d_whitening} for Partial Regression in the two-dimensional case, where the Bayes-optimal weight vector is
\begin{align*}
    w_1^{PR} = \frac{2}{\sqrt{1-a^2\left(c^2-1\right)}}, \quad
    w_2^{PR} = 0.
\end{align*}
Here, when we regress each feature against one another, the effect of the suppressor is removed. 
This is an interesting result for this two-dimensional case, however the residual of each feature depends on the linear combination of all other features, so when dimensionality increases, this will not hold.

Relating these results back to the XAI-TRIS data, where the background features are less explicit suppressor variables than in this two-dimensional setting; when we attempt, for instance, partial regression whitening, background features still may carry residual variance aligning with the signal direction.
All other whitening methods do not succeed at removing the influence of the suppressor from the Bayes-optimal linear model, and when we look at the empirical results of the XAI-TRIS data this makes sense, particular in the non-linear classification scenarios where higher-order dependencies are present.

\section{Discussion and Implications}
The presented analyses highlights the intricate relationship between data preprocessing techniques, specifically whitening, and the explanation performance provided by various XAI methods across different ML models.
While whitening aims to simplify model training and improve numerical stability, its impacts on XAI interpretability are multifaceted and were the main point of investigation in this paper.

The observed results demonstrate that whitening does not offer a fundamental protection against spurious importance attributions to suppressor variables. Such a general effect could only be expected if the observed features are linear combinations of at most as many independent underlying signal or noise factors as there are features, such as for partial regression whitening in the two-dimensional case. When there are more underlying signals than features, whitening will inevitably need to mix discriminative and non-discriminative signal components into novel features, which could lead to worse explanations. 

It is also important to note that these whitening transforms operate on structures that capture linear relationships, such as the covariance or correlation matrix, or feature-wise partial linear regression. 
This may explain reduced effectiveness in the non-linear XAI-TRIS scenarios.
In non-linear settings, whitening has been shown to still improve gradient dynamics and convergence speed in neural networks, \citep{lecun_efficient_2002, kessy_optimal_2018}, the conditioning of the optimization landscape \citep{coates_analysis_2011}, and 
additionally, some architectures (e.g., ICA-based models, t-SNE clustering, or certain generative models) benefit from decorrelated inputs \citep{hyvarinen_independent_2000}. 

Despite these considerations, whitening did have a positive effect on explanation performance depending on the method used. 
Each technique modifies the data in distinct ways, leading to unique alterations in the interpretability maps generated by the XAI methods.
This is evident by the consistent trend where whitening techniques both lead to a shift from diffused to localized importance patterns (that better match the ground truth as seen in the global absolute-valued importance maps) and produced better quantitative results compared to the correlated background case in which no whitening method was applied.
Specifically, optimal signal preserving whitening and symmetric orthogonalization appear to be the most effective in this context, while Cholesky whitening and partial regression seem to be the least effective among the investigated whitening techniques for the XAI-TRIS experiments. 
For partial regression, this may be explained by the methods' inability to fully decorrelate features. 
Cholesky decomposition, on the other hand, depends on the feature ordering. 
Here we only tested one out of $N!$ possible orderings - leaving the top-left pixel intact while successively transforming pixels while moving from the left to the right edge of the image, row by row. 
This somewhat arbitrary sequential orthogonalization approach clearly demonstrates a clear disadvantage compared to the globally optimal maximal signal preservation achieved by symmetric orthogonalization and optimal signal preserving whitening.

The complexities introduced by suppressor variables in XAI interpretations reinforce the need for careful consideration of background noise and its correlation structures when evaluating the performance of XAI methods.
Future work will investigate methods that go beyond these whitening methods to capture and remove non-linear or higher-order redundancy, and how they impact explanation performance, particularly in the presence of suppressor variables.
XAI methods may also require additional mechanisms to distinguish between true predictors and correlated suppressors to maintain adequate explanation performance.

\bibliographystyle{icml2023}
\bibliography{xai_better}

\clearpage
\appendix

\section{Data Generation}\label{app:data-gen}
Following from \citet{clark_xai_2024}, each dataset consists of images of size $8 \times 8$, formulated as $D = \{(x^{(n)}, y^{(n)})\}_{n=1}^{N}$, containing independent and identically distributed observations $(x^{(n)} \in \mathbb{R}^D, y^{(n)} \in \{0, 1\})_{n=1}^{N}$ with $N = 10000$ and the dimensionality of the feature space $D = 64$. The entities $x^{(n)}$ and $y^{(n)}$ represent instances of the stochastic variables $X$ and $Y$, governed by a joint probability density function $p_{X,Y}(x, y)$. 
In each defined scenario, the instance $x^{(n)}$ is synthesized by integrating a signal pattern $a^{(n)} \in \mathbb{R}^D$, which encapsulates the critical features that constitute the ground truth for a model explanation, with background noise $\eta^{(n)} \in \mathbb{R}^D$. In the analysis, a scenario is also considered where the signal pattern $a^{(n)}$ undergoes a random spatial rigid body transformation (involving translation and rotation of the tetromino) $R^{(n)}: \mathbb{R}^D \rightarrow \mathbb{R}^D$. In all other scenarios, the identity transformation is utilized, such that $R^{(n)} \circ a^{(n)} = a^{(n)}$. The transformed signal and noise components, $(R^{(n)} \circ a^{(n)})$ and $(G \circ \eta^{(n)})$, are horizontally concatenated into matrices $A = \left\{(R^{(1)} \circ a^{(1)}), \ldots, (R^{(N)} \circ a^{(N)})\right\}$ and $E = \left\{(G \circ \eta^{(1)}), \ldots, (G \circ \eta^{(N)})\right\}$. The signal and background components are then normalized by the Frobenius norms of $A$ and $E$: $(R^{(n)} \circ a^{(n)}) \leftarrow (R^{(n)} \circ a^{(n)}) / \|A\|_F$ and $(G \circ \eta^{(n)}) \leftarrow (G \circ \eta^{(n)}) / \|E\|_F$, where the Frobenius norm of a matrix $A$ is defined as $\|A\|_F := \left(\sum_{n=1}^{N} \sum_{d=1}^{D} (a_d^{(n)})^2\right)^{1/2}$. Additionally, the weighted sum of the signal and background components is computed, where the scalar parameter $\alpha \in [0, 1]$ determines the SNR.
Two distinct generative models are adopted, diverging based on their method of combining these two elements either additively or multiplicatively. For data generated through either process, each sample $x^{(n)} \in \mathbb{R}^D$ is scaled to the range $[-1, 1]^D$, such that $x^{(n)} \leftarrow x^{(n)} / \max |x^{(n)}|$, where $\max |x^{(n)}|$ denotes the maximum absolute value of the sample $x^{(n)}$.

\subsection{Additive Generation} In scenarios where the model is additive, the data generation formula for the $n$-th sample is defined as:
\begin{equation}
x^{(n)} = \alpha(R^{(n)} \circ a^{(n)}) + (1 - \alpha)(G \circ \eta^{(n)})
\label{eq:add}
\end{equation}
where the signal pattern $a^{(n)} \in \mathbb{R}^D$ varies, embodying tetromino shapes based on the binary class label $y^{(n)}$ which is distributed according to a Bernoulli process with a success probability of 0.5. The noise component $\eta^{(n)}$, indicative of a non-informative background, is derived from a multivariate normal distribution $\mathcal{N}(0, I_D)$, resulting in white Gaussian noise with zero mean and an identity covariance matrix $I_D$. This setup ensures that noise in each feature dimension is independent and follows a standard-normal distribution, designated as the WHITE scenario. In each classification task, an alternate background context, termed CORR, is specified where a two-dimensional Gaussian spatial smoothing filter $G: \mathbb{R}^D \rightarrow \mathbb{R}^D$ modifies the noise element $\eta^{(n)}$, with the smoothing parameter (spatial standard deviation of the Gaussian) set to $\sigma_{\text{smooth}} = 3$.

\subsection{Multiplicative Generation} In scenarios where the model is multiplicative, the sample-wise data generation process is defined as:
\begin{equation}
x^{(n)} = \left(1 - \alpha \left(R^{(n)} \circ a^{(n)}\right)\right) \left(G \circ \eta^{(n)}\right) \label{eq:mult}
\end{equation}
where $a^{(n)}$, $\eta^{(n)}$, $R^{(n)}$, and $G$ are defined as previously stated, with $A$ and $E$ being Frobenius-normalized, and $\mathbf{1} \in \mathbb{R}^D$.
This elaborate approach in generating datasets ensures the creation of a controlled setting crucial for the accurate and systematic assessment of XAI methods. Such an approach also serves to certify that the generated data accurately simulates various realistic scenarios while clearly separating signal from noise, which is pivotal for the analysis and interpretation phases that follow \citep{clark_xai_2024}.

\subsection{Suppressors Emergence}
In the scenarios where background noise is correlated, the presence of suppressor variables is induced in both the additive and the multiplicative data generation cases. A suppressor, in this context, is identified as a pixel not part of the foreground \(R^{(n)} \circ a^{(n)}\), while its activity still finds correlation with a foreground pixel through the application of the smoothing operator \(G\). Drawing on characteristics of suppressor variables previously reported \citep{conger_revised_1974, friedman_graphical_2005, haufe_interpretation_2014, wilming_theoretical_2023}, it is anticipated that XAI methods might erroneously attribute importance to suppressor features in both linear and non-linear settings. This misattribution can lead to decreased explanation performance when compared to scenarios involving white noise backgrounds.

\subsection{Scenarios}
Four distinct types of scenarios are introduced using tetrominoes \citep{golomb_polyominoes_1996}, which are geometric shapes consisting of four features. They are then utilized to define each signal pattern $a^{(n)} \in \mathbb{R}^{8 \times 8}$. Tetrominos were chosen as the basis for signal patterns as they allow a fixed and controllable amount of features (pixels) per sample. Specifically, the T-shaped and L shaped tetrominoes were selected due to their four unique appearances under 90-degree rotations. These tetrominos are used to induce statistical associations between the features and the target in the previously mentioned four different binary classification problems \citep{clark_xai_2024}.

\paragraph{Linear (LIN)} In the linear case, the additive generation model from equation \eqref{eq:add} is employed, where $R^{(n)}$ represents the identity transformation, combining the pure signal pattern and the Gaussian white noise background additively. T-shaped tetromino patterns $a_T$ and L-shaped tetromino patterns $a_L$ are utilized for signal patterns, positioned near the top-left corner if $y = 0$ and near the bottom-right corner if $y = 1$, respectively, thus constituting the binary classification problem. Each four-pixel pattern is encoded such that for each pixel in the tetromino pattern, positioned at the i-th row and j-th column, $a^{T/L}_{i,j} = 1$, and zero otherwise.

\paragraph{Multiplicative (MULT)} The multiplicative generation process \eqref{eq:mult} with signal patterns $a_T$, $a_L$ is defined with the same tetrominoes as in the linear case, while transformation $R^{(n)}$ remains the identity transform. In this scenario, a degree of non-linearity is introduced as the foreground tetromino pattern, when overlaying the background noise, is reduced towards zero. Therefore, values either increase or decrease depending on their original sign. The complexity introduced by the non-linearity renders linear classifiers unable to solve this classification problem effectively \citep{clark_xai_2024}. This configuration is meant to evaluate how different machine learning methods can adjust to and manage intricate, interconnected data presentations that are not linear.

\paragraph{Translations and rotations (RIGID)}
In the RIGID scenario, the defining tetrominoes for each class, denoted as $a^{T/L}$, undergo random translations and rotations. This alteration adheres to a rigid body transform $R^{(n)}$, with the requirement that the entire 4-pixel tetromino must remain within the confines of the image space. Such a constraint ensures that despite the randomness of movement and orientation, the integrity of the tetromino shape is preserved within the visible boundaries of the dataset samples. This process is classified as an additive manipulation, consistent with the guidelines established in equation \eqref{eq:add}. In this context, the complexity introduced by the spatial transformations prevents the effective application of standard linear methods for resolving the classification challenges presented. Instead, such intricate scenarios often necessitate the usage of more sophisticated solutions, typically involving specialized neural network architectures such as Convolutional Neural Networks (CNNs). These architectures are specifically engineered to address the challenges posed by spatial variations within image data, making them better suited for capturing and interpreting the nuanced shifts and rotations applied to the tetromino shapes within the RIGID framework.

\paragraph{Exclusive or (XOR)} 
In the XOR configuration, an additive challenge is presented where both tetromino variants, denoted as $a^{T/L}$, are utilized in each sample, with the transformation $R^{(n)}$ maintaining its role as the identity transform. Within this setup, the class membership is defined such that for the first class (where $y = 0$), a combination of both tetromino shapes is superimposed on the image background, either in a positive or negative overlay, expressed as $a^{XOR++} = a^T + a^L$ and $a^{XOR--} = -a^T - a^L$. Conversely, for the second class (where $y = 1$), the tetromino shapes are displayed in a contrasting manner; one shape is overlaid positively, and the other negatively, denoted as $a^{XOR+-} = a^T - a^L$ and $a^{XOR-+} = -a^T + a^L$. This ensures that all four XOR configurations are represented with equal frequency within the dataset.

\section{Whitening Techniques}\label{app:whitening}

\subsection{Mathematics of Whitening}
\label{subsec:whiteningMath}
Whitening represents a linear transformation applied to a $D$-dimensional random vector $\mathbf{x} = (x_1, \ldots, x_D)^\top$, which has a mean $\mathbb{E}[\mathbf{x}] = \boldsymbol{\mu} = (\mu_1, \ldots, \mu_d)^\top$ and a positive definite $d \times d$ covariance matrix $\text{var}[\mathbf{x}] = \boldsymbol{\Sigma}$. This transformation maps $\mathbf{x}$ to a new random vector:
\begin{equation}
    \mathbf{z} = (z_1, \ldots, z_d)^\top = \mathbf{W}\mathbf{x}
\end{equation}
where $\mathbf{z}$ maintains the same dimension $d$ and has a "white" covariance with unit diagonal, $\text{var}[\mathbf{z}] = \mathbf{I}$. The $d \times d$ matrix $\mathbf{W}$ is termed the whitening matrix. Whitening is especially critical in multivariate data analysis for both computational and statistical simplification and is frequently utilized in preprocessing and as part of modeling \citep{zuber_gene_2009, hao_sparsifying_2015}. Whitening extends beyond merely standardizing a random variable, which is performed through:
\begin{equation}
\mathbf{z} = \mathbf{V}^{-\frac{1}{2}}\mathbf{x}
\end{equation}
with $\mathbf{V} = \text{diag}(\sigma_1^2, \ldots, \sigma_d^2)$ containing the variances $\text{var}[x_i] = \sigma_i^2$. This leads to $\text{var}[z_i] = 1$, although it does not address correlations. Standardization and whitening transformations are often coupled with mean-centering of $\mathbf{x}$ or $\mathbf{z}$ to ensure $\mathbb{E}[\mathbf{z}] = 0$, though this is not mandatory for ensuring unit variances or white covariance. The whitening transformation as defined requires selecting a suitable whitening matrix $\mathbf{W}$. Since $\text{var}[\mathbf{z}] = \mathbf{I}$, it follows that $\mathbf{W}\boldsymbol{\Sigma}\mathbf{W}^\top = \mathbf{I}$, thus $\mathbf{W} (\boldsymbol{\Sigma} \mathbf{W}^{\top} \mathbf{W}) = \mathbf{W}
$, under the condition that
\begin{equation}
    \mathbf{W}^\top\mathbf{W} = \boldsymbol{\Sigma}^{-1}.
\end{equation}
Nevertheless, this condition does not uniquely specify the whitening matrix $\mathbf{W}$. In fact, given $\boldsymbol{\Sigma}$, there are infinitely many matrices $\mathbf{W}$ that fulfill this condition, each leading to a distinct whitening transformation producing orthogonal yet differently sphered random variables \citep{kessy_optimal_2018}.

For all whitening techniques, we regularize the covariance (or correlation, in the case of Optimal Signal Preserving) matrices before further calculation. This is done by checking if the smallest eigenvalue is negative or close to zero (under a threshold of $1 \times 10^{-16}$), and then adding a regularizing value slightly larger than the absolute minimal eigenvalue to the diagonal values of the covariance matrix.

\subsection{Eigendecomposition-based whitening methods} \label{subsec:eigenwhitening}

Each eigendecomposition-based whitening method corresponds to a specific choice of reference matrix $\mathbf{M}$ (either the covariance, overlap, or correlation matrix), which is decomposed as:
\begin{equation}
\mathbf{M} = \mathbf{U} \mathbf{D} \mathbf{U}^\top,
\end{equation}
where $\mathbf{U}$ contains the eigenvectors of $\mathbf{M}$ and $\mathbf{D}$ is the diagonal matrix of eigenvalues.

The whitening transformation is then given by:
\begin{equation}
\mathbf{W} = \mathbf{U} \mathbf{D}^{-1/2} \mathbf{U}^\top.
\end{equation}
This general form yields distinct whitening transformations depending on the matrix $\mathbf{M}$:
\begin{itemize}
    \item \textbf{Sphering (ZCA whitening)}: $\mathbf{M} = \boldsymbol{\Sigma}$ (centered covariance matrix) \cite{kessy_optimal_2018}
    \item \textbf{Symmetric Orthogonalization}: $\mathbf{M} = \boldsymbol{\Sigma} + \boldsymbol{\mu} \boldsymbol{\mu}^{\top}$ (overlap matrix) \citep{annavarapu_singular_2013, colclough_symmetric_2015}
    \item \textbf{Optimal Signal Preservation (OSP)}: $\mathbf{M} = \boldsymbol{\Sigma} \oslash \sigma_\mathbf{X} \sigma_\mathbf{X}^\top$ (correlation matrix), where $\oslash$ is the element-wise division operator \citep{kessy_optimal_2018}
\end{itemize}

The choice of $\mathbf{M}$ produces different whitening transformations while preserving the fundamental structure of symmetric eigenvalue-based decorrelation.

\subsection{Cholesky Whitening}
Cholesky whitening utilizes the Cholesky decomposition to transform a dataset into one where all features are uncorrelated and possess unit variance. This technique ensures that the transformed features have a simpler structure, facilitating more stable numerical computations. The Cholesky whitening procedure encompasses the following steps:
\begin{enumerate}
    \item Compute the covariance of the data matrix $\boldsymbol{\Sigma}$ 
    \item Perform Cholesky decomposition on $\boldsymbol{\Sigma}$, which results in:
        \begin{equation}
            \boldsymbol{\Sigma} = \mathbf{LL}^\top
        \end{equation}
    Here, $\mathbf{L}$ is a lower triangular matrix with real and positive diagonal entries. 
    \item Apply the whitening transformation to obtain the decorrelated feature matrix $\mathbf{X}^{\text{white}}$, computed as:
    \begin{equation}
        \mathbf{X}^{\text{white}} = \mathbf{L}^{-1}\mathbf{X}
    \end{equation}
    where $\mathbf{X}$ denotes the data matrix.
\end{enumerate}
The utilization of the Cholesky whitening matrix leads to the formation of both a cross-covariance matrix and a cross-correlation matrix. These matrices are distinctive for being lower-triangular with positive diagonal elements \citep{kessy_optimal_2018}. The adoption of Cholesky factorization for whitening purposes inherently implies a specific ordering of the variables involved. This ordering is particularly beneficial for time series analysis, as it facilitates the incorporation of auto-correlation effects as highlighted by \citet{pourahmadi_covariance_2011}. The Cholesky whitening process is also recognized for its computational efficiency. Compared to alternative methods such as eigenvalue or singular value decompositions, Cholesky decomposition is generally quicker due to its simpler computational requirements.

\subsection{Partial Regression Whitening}
While not a whitening technique in the above sense, the primary objective of using partial regression as a whitening-like method is to modify each independent variable to isolate its unique variance, minimizing the influence of other variables.
The procedure begins by first centering the data, which is an important step for ensuring that each variable contributes equally to the analysis by removing mean bias.
Then follows the iterative residual calculation step which aims to reduce the influence of other features on each target feature, thereby whitening the dataset:

\begin{enumerate}
    \item For each feature, separate it as the target (to be considered as a temporary dependent variable) from the matrix of remaining features (treated as independent variables), such that $\tilde{\mathbf{y}}_d = \mathbf{x}_{N,d}$ for the $N$ sample values in the $d$-th feature of the $N \times D$ data matrix $\mathbf{X}$.
    \item Taking $\tilde{\mathbf{X}}_d = [\mathbf{x}_{N,0}, \ldots, \mathbf{x}_{N,d-1}, \mathbf{x}_{N,d+1}, \ldots, \mathbf{x}_{N,D} ]$, perform the regression
    \begin{equation}
        \tilde{\mathbf{y}}_d = \tilde{\mathbf{X}}_d \boldsymbol{\beta}_d.
    \end{equation}
    \item Compute regression weights by applying the pseudo-inverse $\tilde{\mathbf{X}}^+_d$ of the matrix of independent variables $\tilde{\mathbf{X}}_d$ to the target feature
    \begin{equation}
        \boldsymbol{\beta}_d = \tilde{\mathbf{X}}^+_d \tilde{\mathbf{y}}_d.
    \end{equation}
    \item Calculate and extract the residuals, which are the portions of the target feature not explained by its linear relationship with the other features. These residuals represent the ``whitened'' features, such that
    \begin{equation}
        \mathbf{X}^{\text{white}}_d = \tilde{\mathbf{y}}_d - \tilde{\mathbf{X}}_d \boldsymbol{\beta}_d.
    \end{equation}
\end{enumerate}

\section{Defining Ground-Truth Feature Importance}\label{app:gt-feature-importance}
Ground truth feature importance is quantitatively defined through the identification of significant pixels, where the significance of a pixel is determined by its statistical relationship with the target outcome \citep{wilming_theoretical_2023}. This leads to the establishment of ground truth sets for significant pixels, considering the positions occupied by tetromino patterns within the dataset, formalized as:

\begin{equation}
    F^+(x^{(n)}) := \{d | \left(R^{(n)} \circ a^{(n)}\right)_d \neq 0, d \in \{1, \ldots, 64\}\}.
    \label{eq:featureIMP}
\end{equation}

In the contexts of both LIN and MULT, each dataset sample includes either a T or an L shaped tetromino, each anchored at predetermined positions, corresponding respectively to the patterns $a_T$ and $a_L$. This structured approach ensures that the absence of a tetromino shape at one specific location is considered as informative as the presence of the alternate shape in a different location, enhancing the comprehensive nature of the pixel importance set in these contexts as:

\begin{equation}
    F^+(x^{(n)}) := \{d | (H \circ a_T)_d \neq 0 \vee (H \circ a_L)_d \neq 0, d \in \{1, \ldots, 64\}\}.
\end{equation}

This conceptual framework is identical to equation \ref{eq:featureIMP} for the XOR challenge and adheres to the operational definition of feature importance as established by \citet{wilming_theoretical_2023}, applied uniformly across the LIN, MULT, and XOR scenarios. In these analyzes, a feature is recognized as significant if it demonstrates a statistical relationship with the target outcome across the dataset under review. Consequently, the most important criterion for any optimal explanation method within this framework is to assign significance exclusively to elements within the set $F^+(x^{(n)})$, thereby ensuring that the attribution of importance is directly tied to statistically relevant features \citep{clark_xai_2024}.

\section{Classifiers}\label{app:classifiers}
Convolutional layers in the CNN architecture are defined with parameters set to enable comprehensive feature analysis: four filters, a kernel size of two, a stride of one, and padding designed to preserve the dimensional integrity between input and output shapes. This padding not only enhances pixel utilization throughout each convolution but also serves to prevent the reduction of output sizes from the already compact images by introducing zero-value filler pixels at the peripheries \citep{clark_xai_2024}. Some widely recognized CNN features like batch normalization are omitted due to compatibility issues with various XAI methodologies. 
For the parameterization $\theta$ and the training dataset $D_{\text{train}}$, classifiers denoted as $f_{\theta} : \mathbb{R}^D \rightarrow Y$ are trained. The training of each network spans over 500 epochs, utilizing the Adam optimizer without regularization. A distinct learning rate is applied based on the scenario: 0.004 for the LIN, MULT, and XOR scenarios, and a reduced rate of 0.0004 for the RIGID scenario to account for its increased complexity. During training, the validation dataset $D_{\text{val}}$ plays a crucial role at each epoch, offering insights into the model's generalization capabilities on unseen data. The validation loss, computed at every epoch, serves as a marker for assessing when the classifier has attained its optimal performance. This is determined by recording the model state at the epoch where the validation loss is at its minimum, a strategy that aids in circumventing typical issues of model overfitting. Upon concluding the training phase, the test dataset $D_{\text{test}}$ is employed to evaluate the finalized model's performance, which is also pivotal in the subsequent analysis of XAI methodologies. A classifier is considered to have effectively generalized the classification challenge if it achieves a test accuracy that meets or surpasses an 80\% threshold.
To accommodate experimentation across a diverse array of XAI methods, each network is constructed within both PyTorch and Keras environments, leveraging a TensorFlow backend. This dual-implementation approach allows for compatibility with a wide range of XAI tools, including those supported by the Captum \citep{kokhlikyan_captum_2020} and iNNvestigate \citep{alber_innvestigate_2018} frameworks.

\section{XAI Methods}\label{app:xai-methods}
\begin{table}[ht]
\centering
\caption{Summary of XAI Methods Analyzed as per \citet{clark_xai_2024}}
\label{tab:XAIMethodsSummary}
\renewcommand{\arraystretch}{2}
\resizebox{\textwidth}{!}{%
\begin{tabular}{|l|p{0.5\linewidth}|l|}
\hline
\textbf{XAI Method} & \textbf{Description} & \textbf{Reference, Framework, parameterization} \\
\hline
Permutation Feature Importance (PFI) & Measures the change in prediction error of the model after permuting each feature’s value. & \citet{fisher_all_2019}, Captum, Default,\\
\hline
Integrated Gradients & Computes gradients along the path from a baseline input to the input sample and cumulates these through integration to form an explanation. & \citet{sundararajan_axiomatic_2017}, Captum, Default, Zero input baseline \\
\hline
Saliency & Computes the gradients with respect to each input feature. & \citet{simonyan_deep_2014}, Captum, Default \\
\hline
Guided Backpropagation & Computes the gradient of the output with respect to the input but ensures only non-negative gradients of ReLU functions are backpropagated. & \citet{springenberg_striving_2015}, Captum, Default \\
\hline
Guided GradCAM & Computes the element-wise product of guided backpropagation attributions with respect to a class-discriminative localization map in the final convolution layer of a CNN. & \citet{selvaraju_grad-cam_2017}, Captum, Default \\
\hline
Deconvolution & Uses a Deconvolutional network to map features to pixels, ensuring only non-negative gradients of ReLU functions are backpropagated. & \citet{zeiler_visualizing_2014}, Captum, Default \\
\hline
DeepLift & Compares the activation of each neuron to its 'reference activation' and produces an explanation based on this difference. &  \citet{shrikumar_learning_2017}, Captum, Default, Zero input baseline \\
\hline
Shapley Value Sampling & Approximates Shapley values by repeatedly sampling random permutations of input features and calculating the contribution of each feature to the prediction. & \citet{castro_polynomial_2009}, Captum, Default, Zero input baseline \\
\hline
Gradient SHAP & Approximates Shapley values by computing the expected values of gradients when randomly sampled from the distribution of baseline samples. & \citet{lundberg_unified_2017}, Captum, Default, Zero input baseline \\
\hline
Kernel SHAP & Approximates Shapley values through the use of LIME, setting the loss function weighting kernel and regularization term in accordance with the SHAP framework. & \citet{lundberg_unified_2017}, Captum, Default, Zero input baseline \\
\hline
Deep SHAP & Approximates Shapley values through the use of DeepLift, computing the DeepLift attribution for each input sample with respect to each baseline sample. & \citet{lundberg_unified_2017}, Captum, Default, Zero input baseline \\
\hline
Locally-interpretable Model Agnostic Explanations (LIME) & Learns a linear surrogate model locally to an individual prediction, perturbing and weighting the dataset in the process, then builds an explanation by interpreting this local model. & \citet{ribeiro__2016}, Captum, Default \\
\hline
Layer-wise Relevance Propagation (LRP) & Propagates the model output back through the network as a measure of relevance, decomposing this score for each model layer. & \citet{bach_pixel-wise_2015}, Captum, Default \\
\hline
Deep Taylor Decomposition (DTD) & Applies a Taylor decomposition from a specified root point to approximate the network's sub-functions, building explanations backward from the output to input variables. & \citet{montavon_explaining_2017}, iNNvestigate, Default \\
\hline
PatternNet & Estimates activation patterns per neuron through signal estimator and back-propagates this through the network. & \citet{kindermans_learning_2018}, iNNvestigate, Default \\
\hline
PatternAttribution & utilizes the theory of PatternNet to estimate the root point for Deep Taylor Decomposition and yields the attribution for weight vector and positive activation patterns. &  \citet{kindermans_learning_2018}, iNNvestigate, Default \\
\hline
\end{tabular}
}
\end{table}

\section{Results}

\subsection{Qualitative Results}

\begin{figure}[ht]
    \centering
    \includegraphics[scale=0.08]{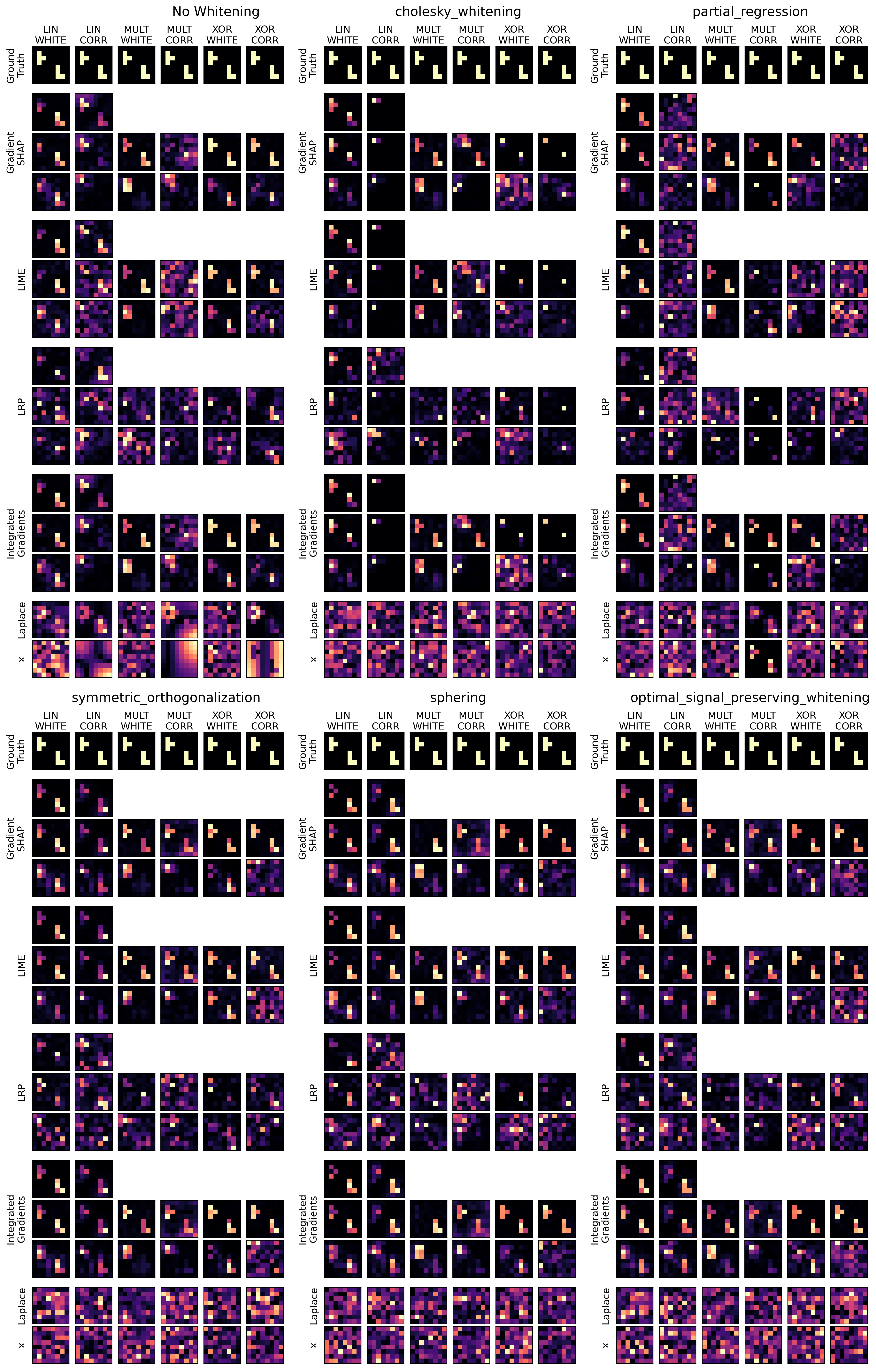}
    \caption{Absolute-valued global importance maps calculated as the mean importance value over all correctly predicted samples, for selected XAI methods and baselines.}
    \label{fig:Gheatmaps}
\end{figure} 

\begin{figure}[ht]
    \centering
    \includegraphics[scale=0.075]{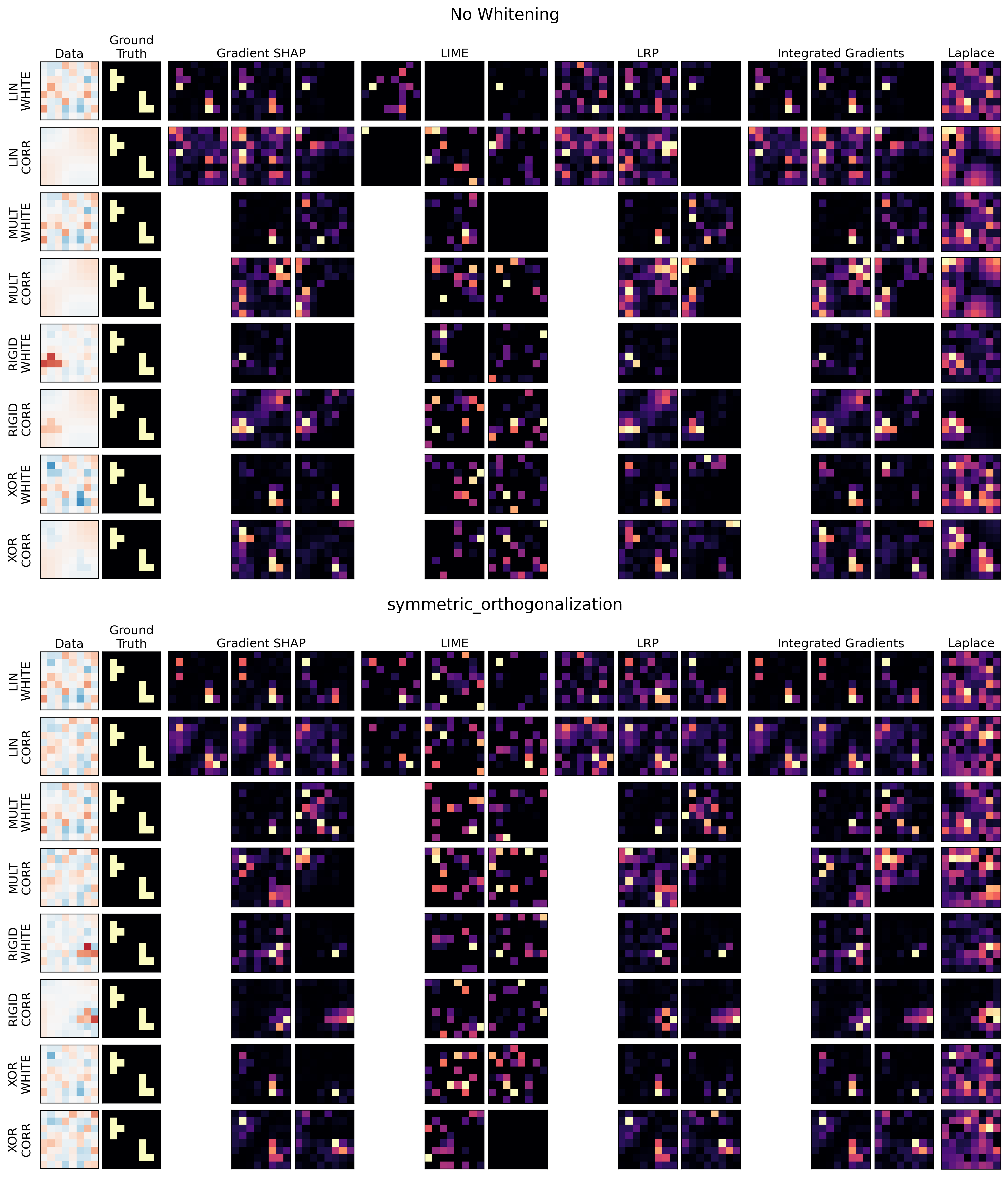}
    \caption{Absolute-valued importance maps obtained for a random correctly-predicted data sample, for data with no whitening applied and data for which the symmetric orthogonalization whitening method was applied. Note, different samples are visualised for both cases.}
    \label{fig:randomSample}
\end{figure}

\subsection{Quantitative Results}

\begin{figure}[ht]
    \centering
    \includegraphics[scale=0.04]{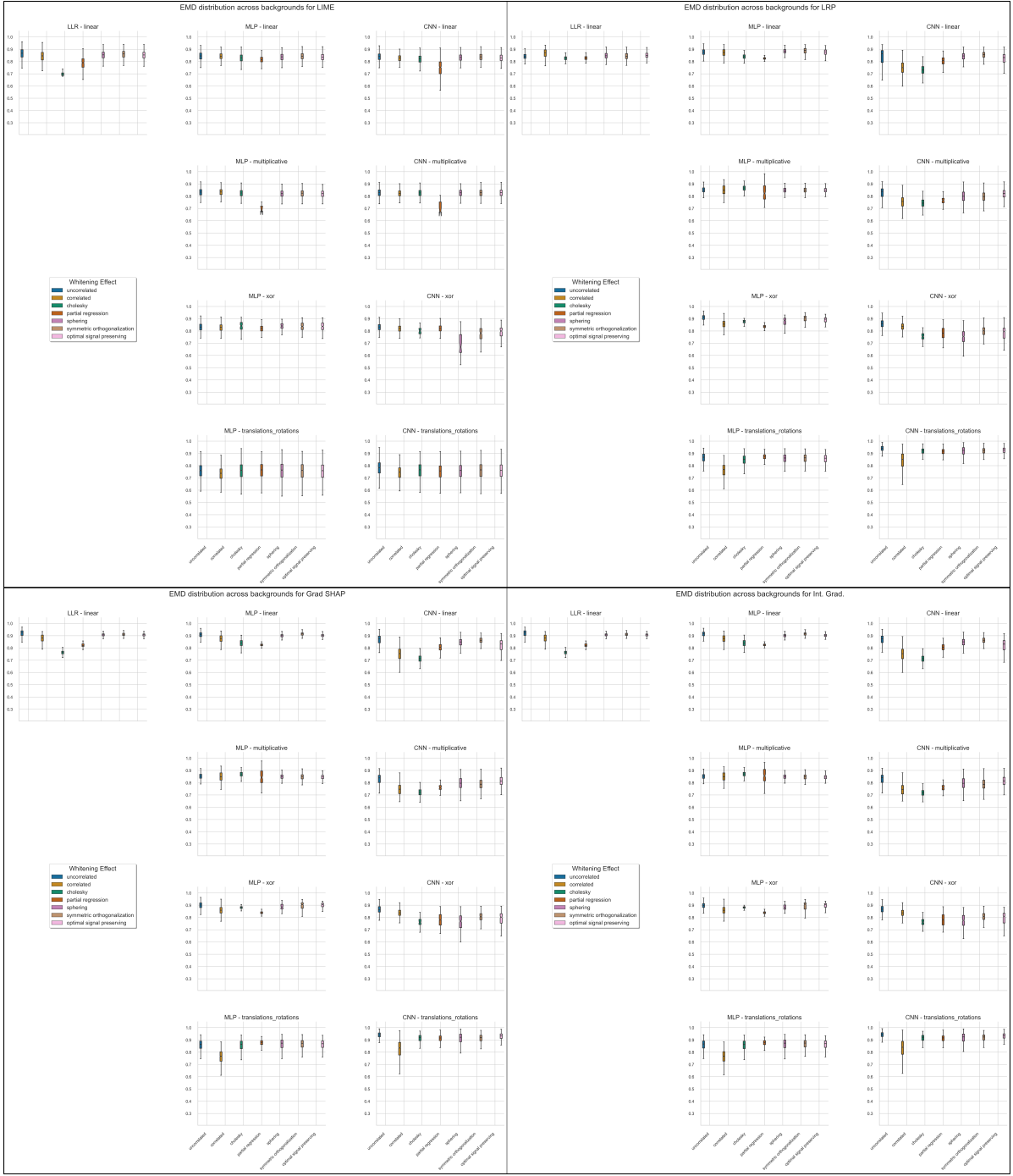}
    \caption{Boxplots of EMD scores across all problem scenarios and background types, where each plot is separated for each of the four main XAI methods studied.}
    \label{fig:EMDFour}
\end{figure}

\begin{figure}[ht]
    \centering
    \includegraphics[scale=0.04]{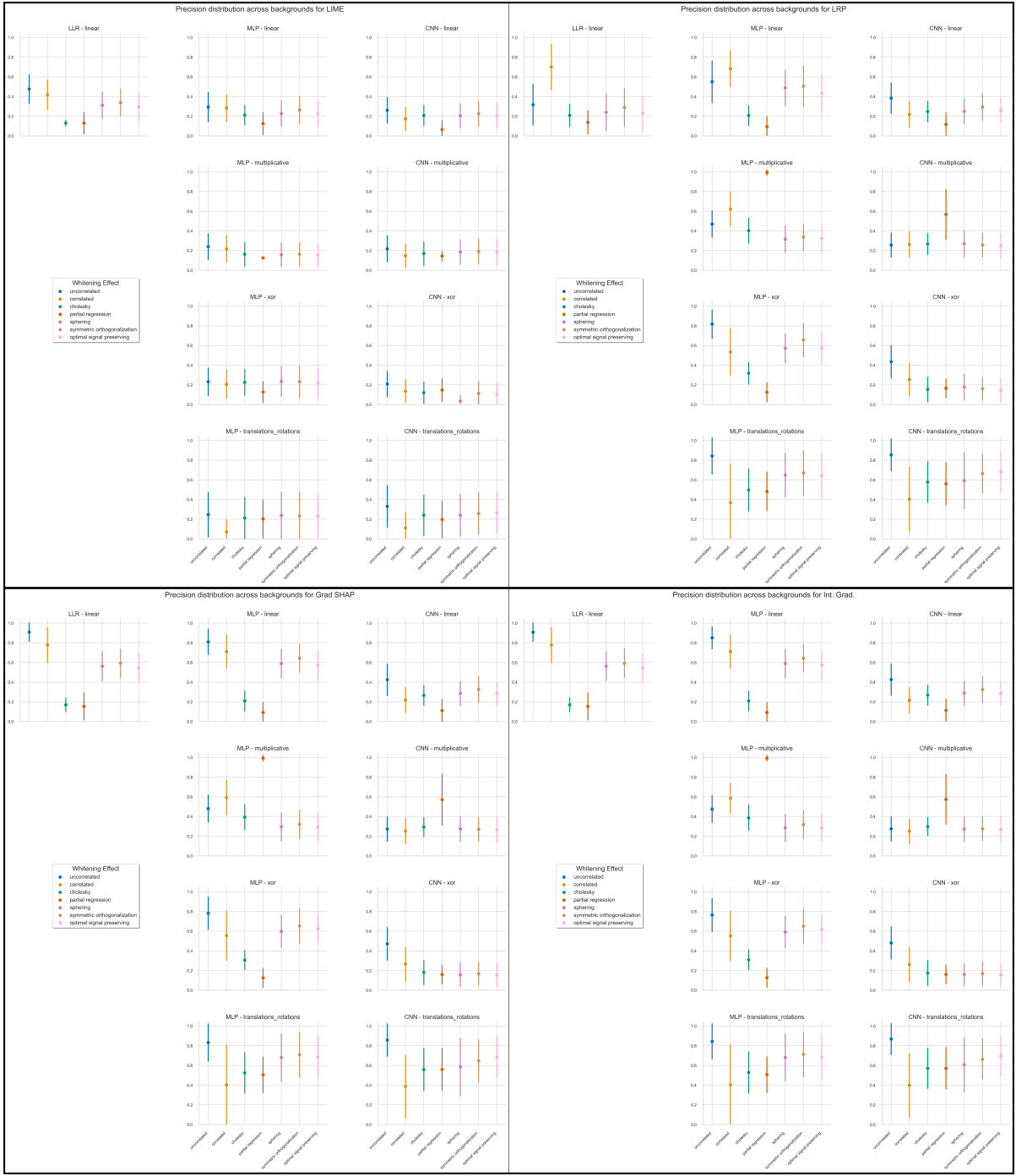}
    \caption{Precision scores across all problem scenarios and background types, where each plot is separated for each of the four main XAI methods studied.}
    \label{fig:precisionFour}
\end{figure}

\begin{figure}[ht]
    \centering
    \includegraphics[scale=0.04]{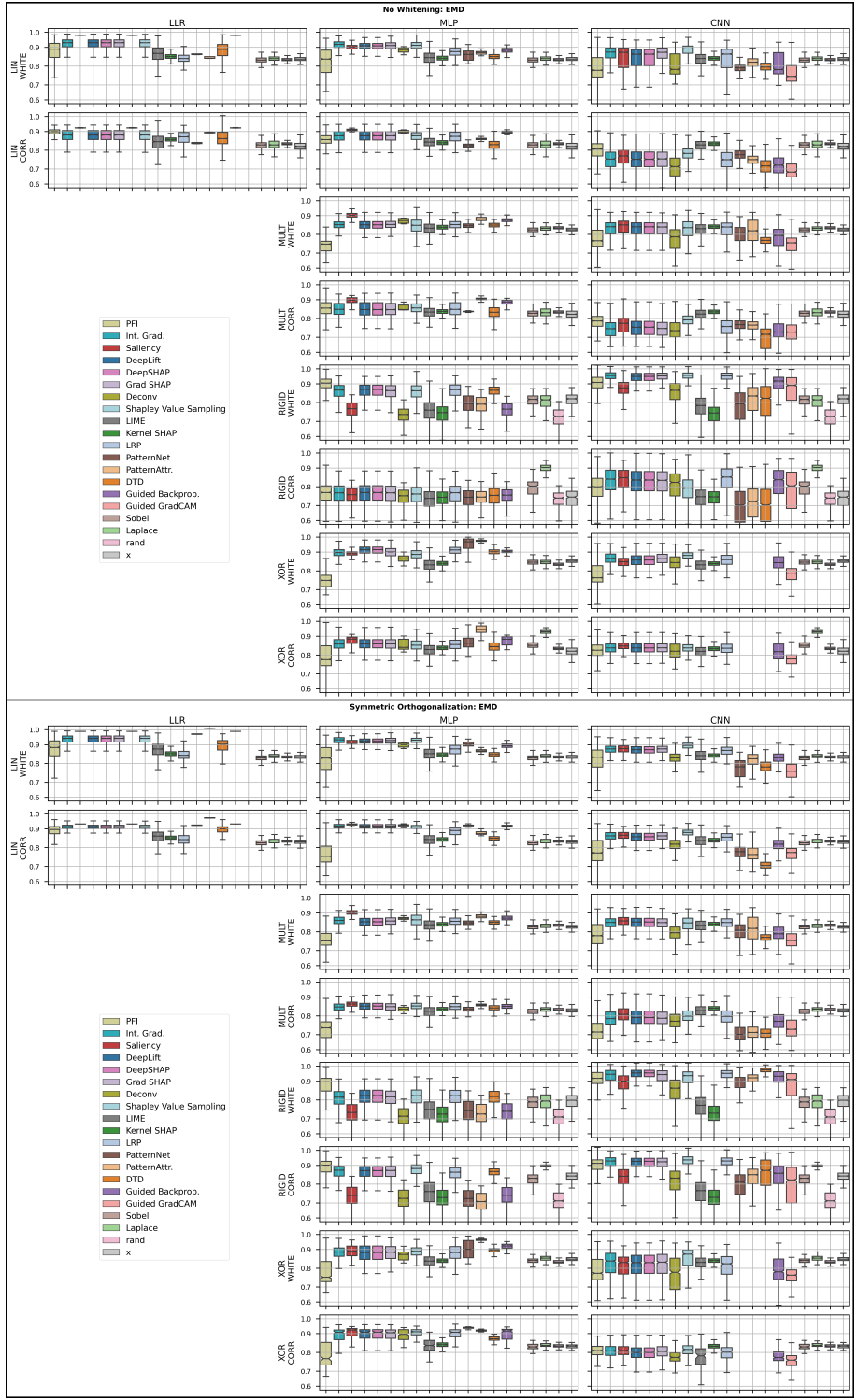}
    \caption{EMD results for all investigated XAI methods and baselines visualised as boxplots of median and quartile scores. The top plot shows the case where no whitening methods are applied to any scenario, and the bottom shows the equivalent where Symmetric Orthogonalization is applied to every scenario, even the WHITE background scenarios. A slight increase in EMD performance can be seen when whitening is applied, whilst retaining the same general trend in XAI method results. }
    \label{fig:methodsEMD}
\end{figure}

\begin{figure}[ht]
    \centering
    \includegraphics[scale=0.04]{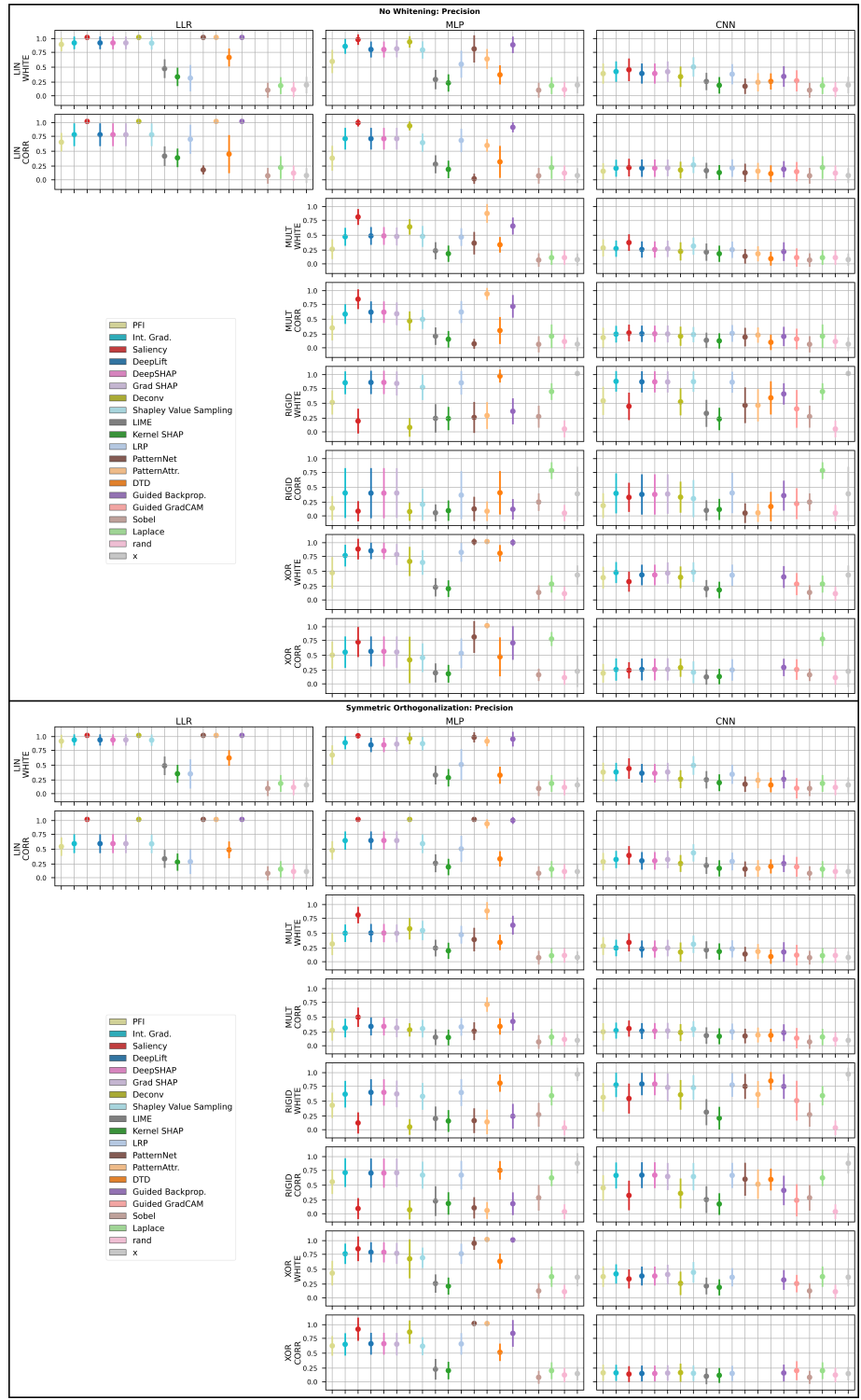}
    \caption{Mean and standard deviation Precision results for all investigated XAI methods and baselines. The top plot shows the case where no whitening methods are applied to any scenario, and the bottom shows the equivalent where Symmetric Orthogonalization is applied to every scenario, even the WHITE background scenarios. A slight increase in EMD performance can be seen when whitening is applied, whilst retaining the same general trend in XAI method results.}
    \label{fig:methodsPrecision}
\end{figure}

\subsection{Theoretical Analysis} \label{app:theory}
For our theoretical analysis, we perform all calculations using Mathematica and then verify the results numerically with our Python implementions of each whitening method studied.

We adopt the two-dimensional linear generative model introduced by \citet{wilming_theoretical_2023}, where $\mathbf{x} = \mathbf{a} z + \eta$ and $y = z$, with $Z \sim Rademacher(1/2)$ the random variable of the realization $z$, signal pattern $\mathbf{a}=(1, 0)^{\top}$ and $H \sim N(\mathbf{0}, \Sigma_\eta)$ the random variable of the realization $\eta$. 
The noise covariance and subsequent data covariance matrices are parameterized as follows:
\begin{equation}
    \label{eq:noise-covariance-app}
    \begin{split}
        \Sigma_\eta = \begin{bmatrix}
                 s_1^2 & c s_1 s_2 \\ c s_1 s_2 & s_2^2 \;,
        \end{bmatrix} \;,
        \quad \Sigma_\mathbf{x} = \begin{bmatrix}
                 s_1^2 + 1& c s_1 s_2 \\ c s_1 s_2 & s_2^2 \;,
        \end{bmatrix} \;,
    \end{split}
\end{equation}

Here, we calculate the whitening transformation matrix $\mathbf{W}$ for the given method, apply it to the data, calculating the whitened covariance matrix $\boldsymbol{\Sigma}_{white} = \mathbf{W}\boldsymbol{\Sigma}\mathbf{W}^{\top}$ and then derive the weights for the the Bayes-optimal linear classifier $f(x) = w^{\top}x + b$. These weights are defined as $w^{\top}=(w_1, w_2)^{\top} = \boldsymbol{\Sigma}_{white}^{-1} (\mu_1 - \mu_2)$ for class means $\mu_1 = (1,0)^{\top}$ and $\mu_2 = (1,0)^{\top}$. 

\subsubsection{Sphering}

Given the size of the data, we can calculate $\boldsymbol{\Sigma}^{-1/2}$ explicitly and do not actually need to manually do the eigendecomposition steps specified in supplementary Section \ref{subsec:eigenwhitening}. With data covariance matrix \( \boldsymbol{\Sigma} \), the Sphering (or ZCA) whitening matrix is calculated as 
\begin{equation}
    \mathbf{W}_{\text{sphering}} = \boldsymbol{\Sigma}^{-1/2}.
\end{equation}
Taking $\alpha = \sqrt{4 s_1^2 s_2^2 c^2 + (s_1^2 - s_2^2 + 1)^2}$,  $\beta  = s_1^2 + s_2^2 + 1$, and $\gamma = \sqrt{8 s_1^2 s_2^2 c^2 + 2(s_1^2 - s_2^2 + 1)^2}$

\[
\mathbf{W}_{sphering} =
\begin{pmatrix}
\displaystyle \frac{\frac{\alpha + s_1^2 - s_2^2 + 1}{\sqrt{\alpha + \beta}} + \frac{\alpha - s_1^2 + s_2^2 - 1}{\sqrt{-\alpha + \beta}}}{\gamma}
&
\displaystyle \frac{s_1 c \left( \sqrt{-\alpha + \beta} - \sqrt{\alpha + \beta} \right)}{\sqrt{2} \cdot \sqrt{ -\left( (s_1^2(c^2 - 1) - 1)(\alpha^2) \right) }}
\\[1.2em]
\displaystyle \frac{s_1 c \left( \sqrt{-\alpha + \beta} - \sqrt{\alpha + \beta} \right)}{\sqrt{2} \cdot \sqrt{ -\left( (s_1^2(c^2 - 1) - 1)(\alpha^2) \right) }}
&
\displaystyle \frac{\frac{\alpha + s_1^2 - s_2^2 + 1}{\sqrt{-\alpha + \beta}} + \frac{\alpha - s_1^2 + s_2^2 - 1}{\sqrt{\alpha + \beta}}}{\gamma}
\end{pmatrix}
\]

\subsubsection{Symmetric Orthogonalization}
Given overlap matrix $\mathbf{M} = \boldsymbol{\Sigma} + \boldsymbol{\mu} \boldsymbol{\mu}^{\top}$ we can similarly define the symmetric orthogonalization whitening matrix as 
\begin{equation}
    \mathbf{W}_{\text{sym}} = \mathbf{M}^{-1/2}
\end{equation}
without carrying out the eigendecomposition explicitly. 
Our python implementation actually does not use the overlap matrix explicitly, however symmetric orthogonalization has been described as such in the aforementioned text to show the similarities to other whitening methods.
Here, we show that the whitening we do is still a valid symmetric orthogonalization.

We begin with the unnormalized scatter matrix:
\begin{equation}
    \mathbf{D} = (N - 1) \cdot \boldsymbol{\Sigma}
\end{equation}

Next, we define a diagonal scaling matrix \( \mathbf{D}_{\text{scaled}} \) containing the square roots of the diagonal elements of \( \mathbf{D} \), i.e., the un-normalized variances:
\begin{equation}
    \mathbf{D}_{\text{ƒ}} = \text{diag}\left( \sqrt{D_{11}},\ \sqrt{D_{22}},\ \ldots,\ \sqrt{D_{dd}} \right) = \text{diag}\left( \sqrt{(N-1)\cdot\Sigma_{11}},\ \sqrt{(N-1)\cdot\Sigma_{22}},\ \ldots \right)
\end{equation}

A scaled version of the scatter matrix is then constructed as:
\begin{equation}
    \mathbf{S}_{\text{scaled}} = \mathbf{D}_{\text{scaled}} \cdot \mathbf{D} \cdot \mathbf{D}_{\text{scaled}} = \mathbf{D}_{\text{scaled}} \cdot (N - 1) \cdot \boldsymbol{\Sigma} \cdot \mathbf{D}_{\text{scaled}}
\end{equation}

This matrix can be interpreted as a diagonally-scaled overlap matrix, which undergoes an eigenvalue decomposition:
\begin{equation}
    \mathbf{S}_{\text{scaled}} = \mathbf{U} \mathbf{\Lambda} \mathbf{U}^\top
\end{equation}

The whitening transformation is then constructed as:
\begin{equation}
    \mathbf{W}_{sym} = \mathbf{D}_{\text{scaled}} \cdot \mathbf{S}_{\text{scaled}}^{-\frac{1}{2}} \cdot \mathbf{D}_{\text{scaled}}
\end{equation}

\begin{figure}[ht!]
    \centering
    \includegraphics[width=0.9\textwidth]{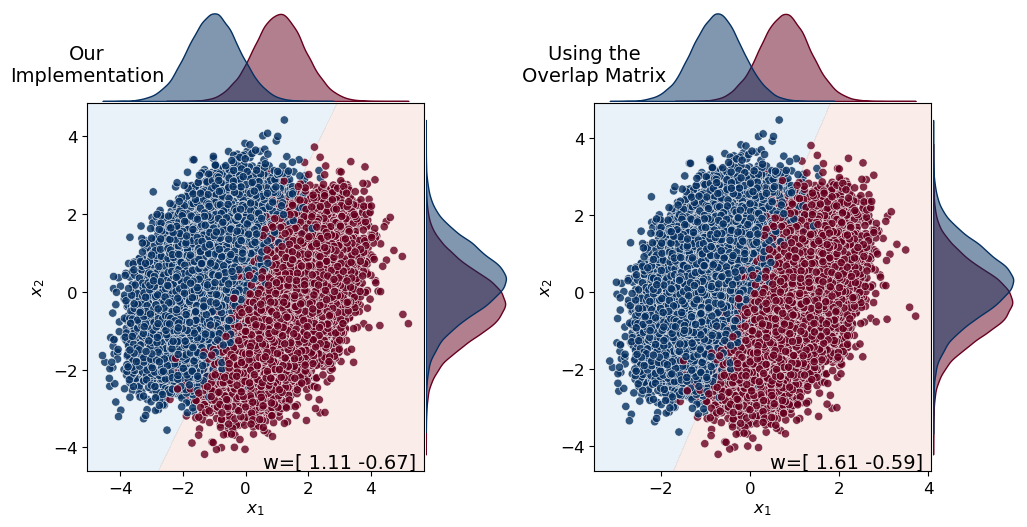}
    \caption{Symmetric orthogonalization whitening of the two-dimensional suppressor data by our formulation (left), and by the formulation explicitly using the overlap matrix (right). Visually the data is identical, and differs only in the scaling across $x_1$. This leads to a different weight matrix, where the difference in weight on $x_2$ is minimal.}
    \label{fig:symm-orth-diffs}
\end{figure}

Even though this construction does not explicitly use the overlap matrix, it produces an equivalent transformation of data to the alternate definition of symmetric orthogonalization whitening, which can be seen in Figure \ref{fig:symm-orth-diffs}.

With $\alpha = \sqrt{4s_1^2 s_2^2 c^2 + (s_1^2 - s_2^2 + 1)^2}$, $\beta  = s_1^2 + s_2^2 + 1$, $\gamma = \sqrt{8 s_1^2 s_2^2 c^2 + 2(s_1^2 - s_2^2 + 1)^2}$, $\delta = \sqrt{-\alpha + \beta}$, and $\epsilon = \sqrt{\alpha + \beta}$, we get

\[
\mathbf{W}_{sym} =
\begin{bmatrix}
\displaystyle
\frac{
\frac{\alpha + s_1^2 - s_2^2 + 1}{\epsilon} +
\frac{\alpha - s_1^2 + s_2^2 - 1}{\delta}
}{\gamma}
&
\displaystyle
\frac{s_1 c (\delta - \epsilon)}{\sqrt{2} \sqrt{-\left((s_1^2(c^2 - 1) - 1) \alpha^2\right)}}
\\[1.2em]
\displaystyle
\frac{s_1 c (\delta - \epsilon)}{\sqrt{2} \sqrt{-\left((s_1^2(c^2 - 1) - 1) \alpha^2\right)}}
&
\displaystyle
\frac{
\frac{\alpha + s_1^2 - s_2^2 + 1}{\delta} +
\frac{\alpha - s_1^2 + s_2^2 - 1}{\epsilon}
}{\gamma}
\end{bmatrix}.
\]

\subsubsection{Optimal Signal Preserving}
Similar to sphering, given the data correlation matrix $Corr(\mathbf{X})=\boldsymbol{\Sigma} \oslash \sigma_\mathbf{X} \sigma_\mathbf{X}^\top$ , the whitening matrix is calculated as 
\begin{equation}
    \mathbf{W}_{OSP} = Corr(\mathbf{X})^{-1/2} = 
\end{equation}

$\alpha = \sqrt{s_1^2 + 1}$
$\beta = \alpha \cdot s_1 c = \sqrt{s_1^2 + 1} \cdot s_1 c$
$\gamma = \sqrt{1 - s_1^2(c^2 - 1)}$

\[
\mathbf{W}_{OSP} =
\begin{bmatrix}
\displaystyle
\frac{ \sqrt{-\beta + s_1^2 + 1} + \sqrt{ \beta + s_1^2 + 1 } }{ 2 \alpha \cdot \gamma }
&
\displaystyle
\frac{ \sqrt{-\beta + s_1^2 + 1} - \sqrt{ \beta + s_1^2 + 1 } }{ 2 s_2 \cdot \gamma }
\\[1.5ex]
\displaystyle
\frac{ \sqrt{1 - \frac{s_1 c}{\alpha}} - \sqrt{ \frac{s_1 c}{\alpha} + 1 } }{ 2 \gamma }
&
\displaystyle
\frac{ \sqrt{ -\beta + s_1^2 + 1 } + \sqrt{ \beta + s_1^2 + 1 } }{ 2 s_2 \cdot \gamma }
\end{bmatrix}
\]

\subsubsection{Cholesky Whitening}
Given the Cholesky decomposition of the Covariance matrix $\boldsymbol{\Sigma} = \mathbf{LL}^\top$, the whitening matrix is simply 

\begin{equation*}
    \mathbf{W}_{\text{Chol}} = \mathbf{L}^{-1} = \left(
\begin{array}{cc}
\displaystyle \frac{1}{\sqrt{s_1^2 + 1}} & 0 \\
\displaystyle -\frac{s_1 c}{\sqrt{(s_1^2 + 1)\left(1 - s_1^2(c^2 - 1)\right)}} & \displaystyle \frac{1}{s_2 \sqrt{1 - \frac{s_1^2 c^2}{s_1^2 + 1}}}
\end{array}
\right)
\end{equation*}

As the ordering of features matters for Cholesky whitening, we also test out permuting the features such that $\mathbf{x} = [x_2, x_1]$, leading to

\begin{align*}
    w_1^{Chol}=-\frac{2ac}{b\sqrt{1-a^2\left(c^2-1\right)}}, \quad w_2^{Chol}=\frac{2}{b}.
\end{align*}

for the weight vector of the Bayes-optimal model.

\subsubsection{Partial Regression}
Given the data covariance matrix:
\[
\boldsymbol{\Sigma} =
\begin{bmatrix}
s_1^2 + 1 & s_1 s_2 c \\
s_1 s_2 c & s_2^2
\end{bmatrix},
\]
we perform partial regression of each variable on the other and compute the corresponding residual variances. The partial regression of \( x_1 \) onto \( x_2 \) is
\begin{align}
\beta_1 &= \frac{\Sigma_{12}}{\Sigma_{22}} = \frac{s_1 s_2 c}{s_2^2}, \\
\text{Var}(x_1 \mid x_2) &= \Sigma_{11} - \beta_1^2 \cdot \Sigma_{22}
= (s_1^2 + 1) - \left( \frac{s_1 s_2 c}{s_2^2} \right)^2 \cdot s_2^2
= (s_1^2 + 1) - \frac{s_1^2 c^2}{s_2^2}.
\end{align}

The corresponding partial regression of \( x_2 \) onto \( x_1 \) is
\begin{align}
\beta_2 &= \frac{\Sigma_{21}}{\Sigma_{11}} = \frac{s_1 s_2 c}{s_1^2 + 1}, \\
\text{Var}(x_2 \mid x_1) &= \Sigma_{22} - \beta_2^2 \cdot \Sigma_{11}
= s_2^2 - \left( \frac{s_1 s_2 c}{s_1^2 + 1} \right)^2 \cdot (s_1^2 + 1)
= s_2^2 - \frac{s_1^2 s_2^2 c^2}{s_1^2 + 1}.
\end{align}

We then form the ``whitening'' matrix \( \mathbf{W}_{\text{PR}} \), which transforms the data to a representation where the residuals from both regressions have unit variance:
\[
\mathbf{W}_{\text{PR}} =
\begin{bmatrix}
\displaystyle \frac{1}{\sqrt{\text{Var}(x_1 \mid x_2)}} &
\displaystyle -\frac{\beta_1}{\sqrt{\text{Var}(x_1 \mid x_2)}} \\[1em]
\displaystyle -\frac{\beta_2}{\sqrt{\text{Var}(x_2 \mid x_1)}} &
\displaystyle \frac{1}{\sqrt{\text{Var}(x_2 \mid x_1)}}
\end{bmatrix}
\]

\end{document}